\pdfoutput=1
\documentclass[11pt]{article}

\usepackage[svgnames]{xcolor}
\usepackage[final]{acl}
\usepackage{verbatim}
\usepackage{times}
\usepackage{latexsym}
\usepackage{amssymb}
\usepackage{booktabs}
\usepackage{multirow}
\usepackage{enumitem}
\usepackage{bm}
\usepackage{makecell}

\usepackage[T1]{fontenc}
\usepackage{tablefootnote}

\usepackage[utf8]{inputenc}
\usepackage{soul}
\usepackage{microtype}
\usepackage{ifsym}

\usepackage{inconsolata}

\usepackage{graphicx}
\usepackage{hyperref}

\colorlet{darkgreen}{green!80!olive!20!}

\newcommand{\longname}{\textbf{Di}versification, \textbf{E}vidence \textbf{T}runcation, and \textbf{Co}mbination for \textbf{K}nowledge-based \textbf{E}lucidation}

\newcommand{\shortname}{\textsc{DietCoke}{}}
\title{Diversify, Rationalize, and Combine: Ensembling Multiple QA Strategies for Zero-shot Knowledge-based VQA}

\author{
  \textbf{Miaoyu Li\textsuperscript{}},
  \textbf{Haoxin Li\textsuperscript{}},
  \textbf{Zilin Du\textsuperscript{}},
  \textbf{and Boyang Li\textsuperscript{}}
\\
  \textsuperscript{}College of Computing and Data Science,\\Nanyang Technological University, Singapore
\\
  {
   \{miaoyu.li, haoxin003, zilin003, boyang.li\}@ntu.edu.sg
  }
}

\begin{document}
\maketitle
\begin{abstract}
Knowledge-based Visual Qustion-answering (K-VQA) often requires the use of background knowledge beyond the image. However, we discover that a single knowledge generation strategy is often insufficient for all K-VQA questions.  To this end, we propose \longname{} (\shortname), which utilizes a bundle of complementary question-answering tactics and aggregates their answers using textual rationales. \shortname{} comprises of three stages: diversification, rationalization, and ensemble. The diversification stage generates three distinctive decision contexts, each leading to its own answer candidate. The rationalization stage generates two rationales, the automatic rationale and the mechanistic rationale, for each answer candidate using decorrelated techniques. Finally, in the ensemble stage, an LLM informed by the rationales selects one answer from the three candidates. Experiments show that \shortname{} significantly outperforms state-of-the-art LLM-based baselines by 2.8\%
on OK-VQA and 4.7\% on A-OKVQA and that the strategies in the ensembles are highly complementary. Code is available at: \href{https://github.com/limiaoyu/DietCoke}{https://github.com/limiaoyu/DietCoke}
\end{abstract}

\section{Introduction}
\begin{figure*}[t]
    \centering
\includegraphics[width=\textwidth]{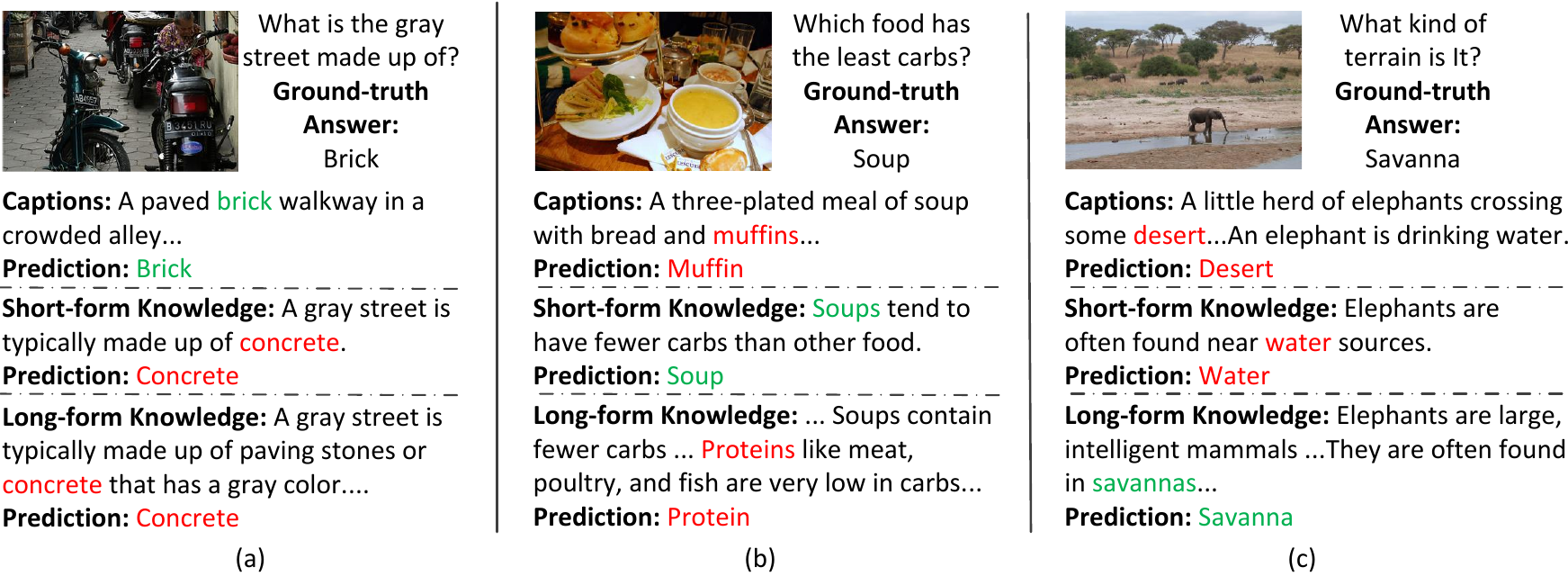}
    \caption{The three K-VQA questions are best answered using distinctive decision contexts, including image captions, (a), and two forms of generated knowledge statements, (b) and (c). This is due to (1) the difficulty in controlling the generation of captions and knowledge statements and ensuring they contain relevant information, and (2) the inability of LLMs to identify correct answers from noisy contexts.}
    \label{fig:intro}
\end{figure*}

Knowledge-based Visual Question-answering (K-VQA) requires background knowledge beyond the image content. For example, to answer the question in Fig. \ref{fig:intro} (b), it is necessary to know the carbohydrate content of different types of food. Zero-shot K-VQA provides an effective evaluation of AI models in applying their knowledge to answer novel vision-informed questions.


An effective technique for zero-shot VQA is to translate an image to a textual decision context for a text-based model, such as an LLM, which the answers the question using the context \citep{pica,pnpvqa}. For example, \citet{pica} translate the image to captions; \citet{img2llm} further generate question-answer pairs to demonstrate the VQA task. To cater to the knowledge-intensive nature of K-VQA, \citet{KGVQA} generate a paragraph describing relevant background knowledge, from which an LLM extracts the answer. These approaches rely on a single question-answering strategy using a single type of decision context.



Interestingly, we observe that one LLM strategy is often not sufficient for K-VQA datasets. For some questions, merely the captions are sufficient for finding the right answer. Other questions benefit from knowledge extracted from LLMs, but it is often not clear if a short, one-sentence knowledge statement or a paragraph would beget the right answer from the LLM. 
As an illustration, in Fig. \ref{fig:intro} we show three questions and three AI-generated decision contexts for each question, including captions, short-form knowledge as a single sentence, and long-form knowledge, which contains multiple sentences. In each question, only one decision context leads to the right answer. 



\sloppy{
In light of this, we propose \longname{} (\shortname) to achieve dynamic ensemble of different question-answering strategies. \shortname{} comprises a diversification phase, a rationalization phase, and an ensemble phase. In each phase, we use an in-context-learning LLM without training. In diversification, we generate three decision contexts for each K-VQA question, including image captions, short-form knowledge, and long-form knowledge, with an increasing amount of background knowledge. The LLM generates three answer candidates, one from each decision context.}


In the ensemble phase, the LLM selects one correct answer from the three candidates. To encourage informed decisions, we design the rationalization phase, which generates two complementary types of rationales for each answer candidate. A rationale summarizes the portion of a decision context that can support the correctness of the answer. The first type, automatic rationale, is generated by simply asking an LLM to summarize the reasoning behind the answer into one sentence. The second type, mechanistic rationale, is one sentence from the decision context that contributes the most to the generation of the answer, as identified by a model interpretation technique, GradCam \citep{gradcam}. These brief rationales preserve the most relevant information in the original decision contexts, and prevent irrelevant information from misleading the ensemble LLM. Informed by the rationales, the LLM in the ensemble phase selects one final answer from the candidates, achieving dynamic ensemble of the question-answering strategies. 

On two popular K-VQA datasets, OKVQA and A-OKVQA,  \shortname{} outperforms state-of-the-art baselines that use frozen LLMs without training by 2.6\% to 4.7\%. Ablation studies reveal that the fusion of three answers improves performance over the best single answer strategy by 1.1\% on OK-VQA and 1.6\% on A-OKVQA. Further, adding both the automatic and mechanistic rationales obtains gains of 1.2\% and 1.4\%, respectively on the two datasets.

The success of \shortname{} can be understood from multiple perspectives. First, \shortname{} creates an effective ensemble by employing decorrelated QA strategies and rationalization strategies. Classic theory on ensemble learning 
\cite{Clemen-1985,Breiman-random-forest-2001} 
indicates that an ensemble of weak classifiers becomes more powerful as the individual classifiers become less correlated and more diverse. Intuitively, different QA contexts lead to diverse answers. Also,  disparate mechanisms used by the two rationalization strategies utilize reduce correlation. In \S \ref{sec:ablation}, we perform a series of ablation studies that demonstrate the synergy among the three QA strategies and the two rationalization strategies. 

Second, rationales in \shortname{} can be understood from a chain-of-thought (CoT) perspective. Recently, \citet{li2024:Chain-of-Thought} prove that chain-of-thought prompting expands the circuit complexity of problems solvable by an LLM from the class $AC^0$ with constant circuit depth to the class $P/poly$ with polynomial circuit depth. This is made possible by saving intermediate computational results as CoT tokens. Analogically, providing the QA contexts (\emph{i.e.,} intermediate results) to the  answer-selecting ensemble LLM should be beneficial. However, in practice we observe that long decision contexts may contain many irrelevant facts that can mislead the ensemble LLM. By summarizing the contexts into rationales, we provide abridged chains-of-thoughts that expand circuit depth while avoiding misleading tokens. 






Our contributions include:
\begin{itemize}[topsep=4pt,itemsep=4pt,partopsep=4pt, parsep=4pt]
\item We identify the need for combined use of multiple question-answering strategies in K-VQA, and propose \shortname{}, which fuses multiple answers and answer strategies dynamically using frozen in-context learning LLMs, achieving state-of-the-art performance on zero-shot OKVQA and A-OKVQA. 
%
%
\item We propose that rationales serve an important role in answer ensemble. Further, we devise two rationale generation techniques, automatic LLM summarization, and mechanistic LLM interpretation. These rationales complement each other in enhancing VQA performance.
\end{itemize}

\section{Related Work}
\sloppy{
\paragraph{Knowledge-based VQA. } Knowledge-based VQA requires background knowledge beyond the image content to answer questions. The popular K-VQA datasets include OK-VQA \citep{okvqa} and A-OKVQA \citep{aokvqa}, both of which cover a wide range of knowledge categories. For OK-VQA, answering questions usually only requires knowledge recall, without the need for additional reasoning. A-OKVQA, on the other hand, is an enhanced version of OK-VQA, where questions often require further reasoning based on background knowledge to answer. Early studies \citep{garderes2020conceptbert-exter1,marino2021krisp-exter2,wu2022multi-exter3,mucko-exter4,outbox-exter5,narasimhan2018straight-exter6,gao2022transform-exter7} retrieve related knowledge from external knowledge bases such as Wikipedia, ConceptNet, and Google Images. Recent approaches \citep{pica,hu2022promptcap-imp2,wang2023filling,shao2023prompting-imp1,xing2023toa-imp3,si2023combo-imp4,decompsition} require no explicit knowledge retrieval; they leverage implicit world knowledge stored in LLMs or Large Vision-Language Models (LVLMs) by using them as QA models. To further improve the QA performance of LLMs or LVLMs, many few-shot methods \citep{pica,shao2023prompting-imp1,xenos2023simple,decompsition} are proposed, which use the training samples as exemplars in the prompt. Unlike existing LLM-based methods that utilize knowledge implicitly, our method explicitly generates relevant background knowledge from an LLM and uses it to assist in answering questions.}


\paragraph{End-to-end LVLMs and LVLM Prompting for Zero-shot K-VQA. } End-to-end training is adopted to align the vision and text modalities, so that a pretrained visual encoder can work seamlessly with a pretrained LLM, resulting in LVLMs \citep{wang2022ofa,Flamingo-train1,PromptFuse-train3,Mapl-train4,blip2,MixPHM-train2,zhu2023minigpt,lin2024moe,liu2024improved}. While these LVLMs achieve high performance on zero-shot K-VQA, they require training on an enormous quantity of image-text pairs and are hence compute-intensive. 

To enable the LVLMs to better understand questions and perform reasoning, one concurrent method explore question rephrasing in the prompt \citep{REPARE2024}. However, it is difficult for this method to change to a different LLM, which provides crucial knowledge for K-VQA, without expensive retraining of the LVLMs. In contrast, our method translates images into textual descriptions and directly utilizes frozen LLMs to answer questions, enabling us to easily change the knowledge source at any time.

\paragraph{Text-mediated Zero-shot K-VQA. } Another type of zero-shot method converts visual information to textual descriptions and applies frozen LLMs to answer questions based on these descriptions alone. We classify these methods based on the number of types of descriptions used and whether answer ensemble is employed. Most methods \citep{pica,pnpvqa,img2llm,cola} translate images into a single type of description: captions, and generate one answer from them. \citet{cola} uses two different LVLMs to generate different captions and answer candidates, and one LLM to choose the final answer. To accommodate the need for external knowledge in K-VQA, \citet{KGVQA} generate two types of descriptions, captions and long-form knowledge and concatenates into one prompt. Hence, we consider it to be a single-strategy method. In contrast, \shortname{} explicitly combines different QA strategies by selecting from their answers. 



\begin{figure*}[t]
\centering
\includegraphics[width=\linewidth]{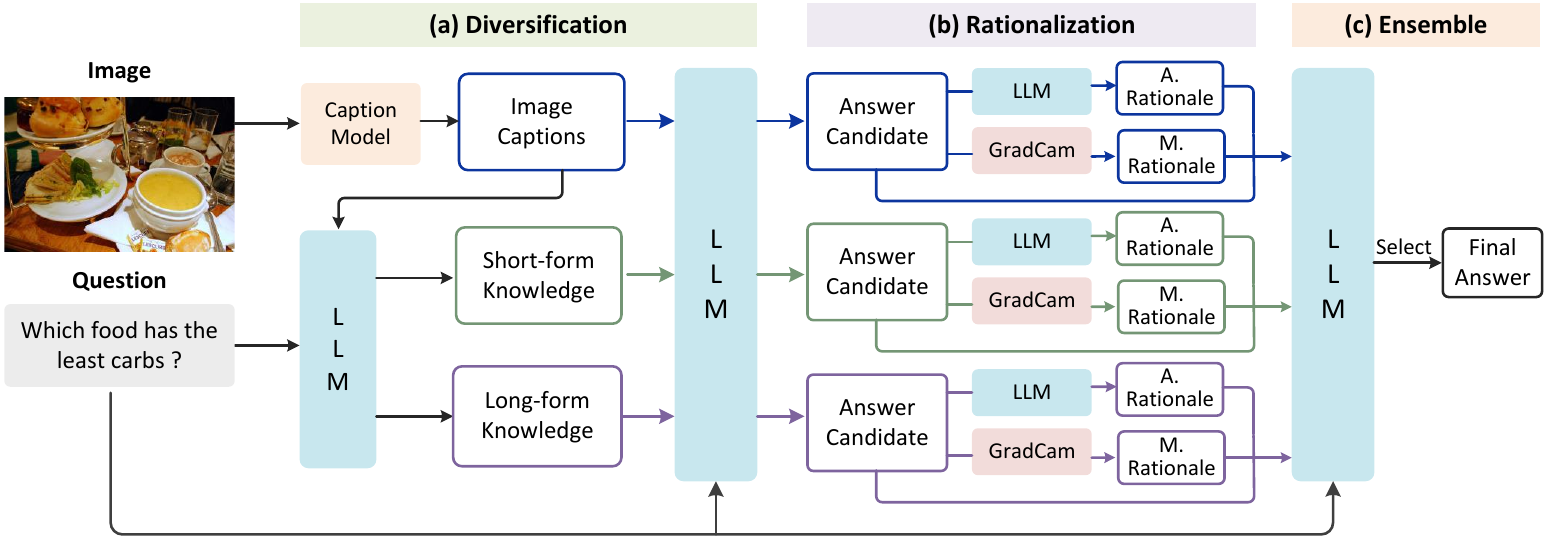}
\caption{The framework of \shortname{}. The diversification phase contains three question-answering strategies, which generate three different decision contexts for the answer-predicting LLM. The short-form knowledge contains a single sentence, whereas the long-form knowledge consists of a paragraph of background information. The caption-only strategy generates only image captions and no background knowledge. For each strategy, the LLM generates an answer candidate. In the rationalization phase, for each candidate, we generate an Automatic (A.) Rationale and a Mechanistic (M.) Rationale, which guide the ensemble LLM in selecting the best answer. } 
\label{fig:framework}
\end{figure*}

\section{Method}

The goal of our method is to jointly utilize multiple decision contexts to answer each question and achieve a dynamic ensemble of these different strategies. To accomplish this, we propose \shortname{}, which consists of three phases: diversification, rationalization and ensemble. This section provides detailed descriptions of them.

\subsection{Diversification}
In diversification phase, we generate three types of decision contexts for each question, including captions, short-form knowledge consisting of one sentence, and long-form knowledge containing multiple sentences.  
They provide three different strategies for each question, resulting in three answer candidates. 

\paragraph{The Caption-only Strategy.}
Since LLMs can only process textual information, we use an off-the-shelf image captioning model to transform an image into multiple textual captions. To ensure the relevance to the question, we employ the question-guided caption generation mechanism in PNP-VQA \citep{pnpvqa}, which first identifies the image patches most relevant to the question and then generates captions only from these relevant patches.
The prompt for the caption-only strategy is shown in Tab. \ref{tab:PromAns} of Appendix \ref{sec:appendix}.

To achieve in-context learning, we follow Img2LLM \citep{img2llm} and generate Question-answer (QA) pairs from the image captions. The QA pairs serve as demonstration of the QA task for LLM in-context learning. As the task demonstration is necessary for LLMs to perform QA, we include these QA pairs in the prompt of \emph{all three} strategies of diversification.



\paragraph{The Short-form Knowledge Strategy.} We generate relevant background knowledge for each question using an LLM. To ensure that the generated knowledge is relevant to the question and the image content, we provide the captions and the question in prompt: ``\textit{$\langle$Captions$\rangle$ $\langle$Question$\rangle$ Please provide background knowledge related to this question in a single sentence.}''. To generate short-form knowledge, we add the length constraint ``\textit{in a single sentence}''. The short-form knowledge ideally represents the most relevant background knowledge to the question. The prompts for generating short-form knowledge and the answer to the question are detailed in Tab. \ref{tab:PromKnow} and Tab. \ref{tab:PromAns} in Appendix \ref{sec:appendix}.

We observe that short-form knowledge tends to be highly relevant to the visual question. However, due to its short length, it may not be comprehensive enough to get the correct answer.

\paragraph{The Long-form Knowledge Strategy. } 
In this strategy, we remove the length constraint ``\textit{in a single sentence}'' from the prompt, so that the LLM can generate as much text as it wants. In most cases, this leads to one or two paragraphs, which ideally could capture comprehensive background knowledge related to the question. The prompts for generating long-form knowledge and utilizing it to generate answer are detailed in Tab. \ref{tab:PromKnow} and Tab. \ref{tab:PromAns} of Appendix \ref{sec:appendix}.

The short-form knowledge and the long-form knowledge are synergistic. 
Constrained to a single sentence, the short-form knowledge context is succinct but may leave out relevant information. In comparison, the long-form knowledge strategy prioritizes knowledge recall over precision, and may contain many irrelevant facts that potentially distract the LLM. Their synergy is further demonstrated in the experiments (\S \ref{sec:ablation} and Fig. \ref{fig:synergy-of-QA-strategies}).  Balancing different strategies is hence important for high performance. 


\subsection{Rationalization}
Since not all question-answering strategies are suitable for a given question, selecting the correct answer is crucial. To allow the LLM to make informed choices, the rationalization phase generates two types of rationales with different methods for each answer candidate, which we call automatic rationales and mechanistic rationales. These brief rationales retain only the most relevant information from the decision context, thus mitigating the risk of the LLM being misled by a plethora of irrelevant information.

\paragraph{Automatic Rationale.}
To generate the automatic rationale, we feed the original question, the decision context, and the predicted answer to an LLM and directly ask the LLM to summarize the rationale behind the answer to a single sentence. The LLM is not restricted to selecting a sentence from the decision context and performs open-ended generation. 
The prompt for generating automatic rationales is shown in Tab. \ref{tab:PromAR} of Appendix \ref{sec:appendix}.

\paragraph{Mechanistic Rationale. } Though the automatic rationale is easy to acquire and often reasonable, there are occasions when the LLM generate incorrect rationales by ignoring part of the decision context or hallucinating (see examples in Fig.~\ref{fig:cases2_appendix}). Although this could be partially alleviated by exhaustive prompt engineering, we opt for a more systematic solution, which is to obtain the rationale through mechanistic interpretation of the answer-generating LLM.

Inspired by GradCAM \citep{gradcam}, we devise a method to compute the contribution of each sentence in the decision context to the answer candidate. We first introduce some notations. The prompt to the LLM, including the instruction and the decision context (captions or knowledge statements), contains $N_P$ tokens. The answer generated by the LLM contains $N_A$ tokens. When generating the $k$-th answer token $w_k$, the Transformer-based LLM attends to all previous $N_k = N_P + k - 1$ tokens. We extract the attention weight vector $\bm{a}^{(h)} \in \mathbb{R}^{N_{k}}$ from the $h$-th attention head. The components of $\bm{a}^{(h)}$ sum up to 1, $\sum_{i=1}^{N_k} a^{(h)}_i = 1$. 


We seek the contribution of every token preceding the $k$-th answer token. Instead of the raw attention scores, which may be redundant and inaccurate, we weigh the attention scores with its gradient from the probability $p_k$ of the predicted answer token $w_k$. We disregard negative gradients as we focus on positive contributions. Formally, the relevance scores of each of the $N_k$ tokens is
\begin{equation}
  \label{eq:token_relevance}
  \bm{r}_k= \frac{1}{H} \sum_{h=1}^{H} \max \left(0,\frac{\partial p_k}{\partial  \bm{a}^{(h)}} \right) \bm{a}^{(h)},
\end{equation}
where $H$ is the number of attention heads. Next, we re-normalize and aggregate the contribution of prompt tokens to all answer tokens. 
\begin{equation}
  \label{eq:answer_relevance}
  \bm{r} = \sum_{k=1}^{N_{A}} \textit{softmax}(\bm{r}_{k}[0:N_{P}]),
\end{equation}
where $[0:N_{P}]$ selects the first $N_P$ components of $\bm{r}_{k}$. 
%
Finally, we obtain sentence-level contributions by summing over the contributions of all tokens in each sentence in the decision context. The sentence with the highest contribution is picked as the mechanistic rationale. As the short-form knowledge context contains only a single sentence, it is always selected as the mechanistic rationale. 


\subsection{Ensemble}
\sloppy{
In the final ensemble stage, we provide all three answer candidates with corresponding rationales, including both the automatic and the mechanistic rationales, in the prompt. The LLM is instructed to select the best answer, thereby achieving dynamic strategy ensemble. To ensure the model perceives the image content during ensemble, we also include caption-generated QA pairs in the prompt. The exact prompt for question-answering strategy fusion is in Tab. \ref{tab:PromFus} of Appendix \ref{sec:appendix}.}

\section{Experiment}

\begin{table*}[t]
\centering
\resizebox{0.8\linewidth}{!}{
\begin{tabular}{@{}l|llcc|ccc@{}}
\toprule
\multirowcell{2}{Method} & \multirowcell{2}{QA\\Model} & \multirowcell{2}{Model\\Size} & \multicolumn{2}{c|}{\makecell{Decision Context}} & OK-VQA & \multicolumn{2}{c}{A-OKVQA} \\
& &  & Captions & Knowledge  & test & val  & test  \\ \midrule
\multicolumn{8}{c}{\textit{LLM-based Zero-shot Methods}} \\
{PICa} & GPT-3 & 175B & \checkmark & $\times$ & 17.7  & -  & -             \\
{PNP-VQA}   & UnifiedQA  & 11B & \checkmark & $\times$ & 35.9                       & -     & -             \\
{Cola-Zero} & FlanT5   & 11B & \checkmark   & $\times$ & 39.4 & -   & - \\
{Img2LLM} & OPT     & 175B & \checkmark  & $\times$ & 45.6  & 42.9  & 40.7 \\
{Img2LLM}$^*$ & Gemma   & 7B  & \checkmark & $\times$ & 45.6$^*$  & 44.9$^*$  & -  \\
{Img2LLM}$^*$ & Mistral  & 7B & \checkmark & $\times$& 46.3$^*$ & 44.3$^*$ & -             \\
{KGenVQA} & UnifiedQA  & 11B  & \checkmark & long  & 45.4   & 39.1  & - \\
{RQprompt} & GPT-3  &175B & \checkmark & $\times$ & 46.4 & 43.2 & 43.9  \\ 
{\shortname}  & Gemma  & 7B &\checkmark & long \& short &  \underline{47.6}  & 47.3    & \underline{46.8}          \\
\shortname & Mistral  & 7B & \checkmark & long \& short & \textbf{49.2} & \underline{47.5}  & \textbf{48.6} \\
\midrule
\multicolumn{8}{c}{\textit{\textcolor{gray}{LVLM-based Zero-shot Methods}}} \\
{\textcolor{gray}{REPARE}} & \textcolor{gray}{BLIP2}  &\textcolor{gray}{3B} & - & - & - & \textcolor{gray}{44.9} & -\\
{\textcolor{gray}{REPARE}} & \textcolor{gray}{BLIP2}  &\textcolor{gray}{11B} & - & - & - & \textcolor{gray}{47.3} & -\\
{\textcolor{gray}{REPARE}} & \textcolor{gray}{MiniGPT-4}  &\textcolor{gray}{7B} & - & - & - & \textcolor{gray}{33.2} & - \\
{\textcolor{gray}{REPARE}} & \textcolor{gray}{MiniGPT-4}  &\textcolor{gray}{13B} & - & - & - & \textbf{\textcolor{gray}{47.9}} & - \\
\bottomrule
\end{tabular}}
\caption{Comparison with state-of-the-art methods on zero-shot K-VQA. We report the VQA score of direct answer on OK-VQA and A-OKVQA datasets. The best score is indicated in \textbf{bold}, while the second best score is indicated in \underline{underline}. The results marked with * represent the baselines we implemented ourselves. }
\label{tab:results}
\end{table*}

\begin{figure}[t]
    \centering
\includegraphics[width=\linewidth]{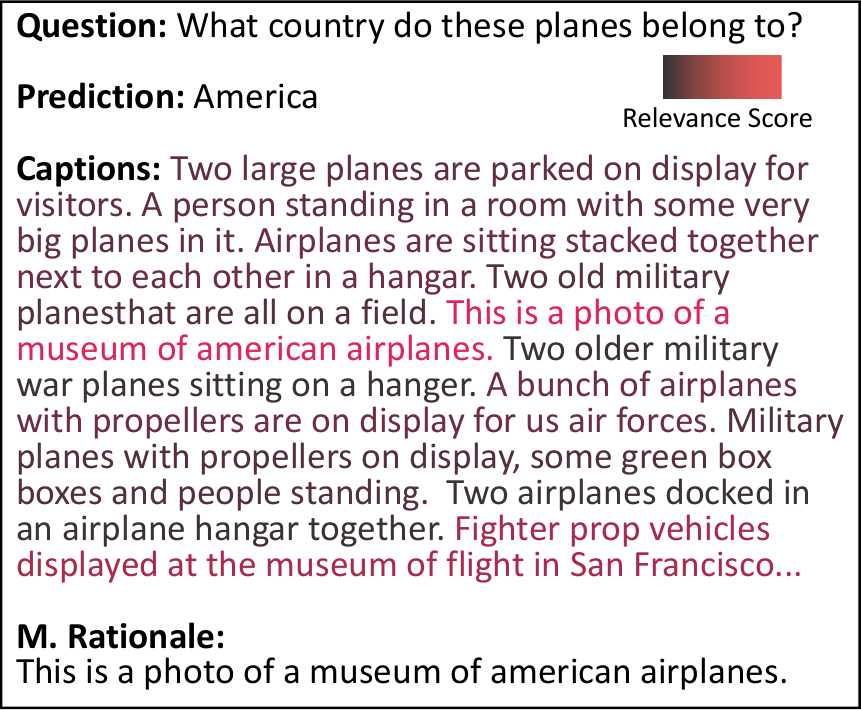}
    \caption{The heatmap of relevance scores of sentences in captions. The relevance score represents the contribution of the sentence to the answer.}
    \label{fig:heatmap}
\end{figure}

\subsection{Setup}
\paragraph{Datasets.} We evaluate our method on two mainstream K-VQA datasets: OK-VQA \citep{okvqa} and A-OKVQA \citep{aokvqa}. Questions in both datasets require knowledge beyond the images to answer. We utilize the test split of OK-VQA and the validation split and the test split of A-OKVQA for evaluation. These splits contain 5,046, 1,100, and 6,700 questions, respectively. We follow the official evaluation protocols of direct answer and report VQA scores for each dataset.

\paragraph{Implementation Details.} Since the quality of captions significantly impacts the results, we follow the previous approaches \citep{img2llm, lan2023improving, KGVQA} to use BLIP \citep{blip} as the captioning model. We generate 30 captions for each image. Additionally, we utilize a finetuned T5-large model \citep{t5}, as in Img2LLM, to generate 30 QA pairs from the captions to enable in-context learning when generating answers. To demonstrate the generalization ability of our method, we use two different models, Mistral-7B and Gemma-7B, in our experiments. The LLMs cannot access answer lists or training samples, achieving zero-shot K-VQA.

\paragraph{Baselines. }
We compare \shortname{} with previous zero-shot K-VQA methods without training. The methods using frozen LLMs can be roughly divided into two categories: (1) Methods employing the caption-only strategy, PICa \citep{pica}, PNP-VQA \citep{pnpvqa}, Cola-Zero \citep{cola}, Img2LLM \citep{img2llm}, and RQprompt \citep{lan2023improving}. (2) Methods employing the long-form knowledge strategy. The only method in this category is KGenVQA \citep{KGVQA}. Unlike translating visual information into textual descriptions for frozen LLMs, the concurrent method REPARE \citep{REPARE2024} directly utilizes an LVLM as QA model\footnote{The REPARE+LLaVA-1.5 model does not meet the zero-shot requirement, as  LLaVA-1.5 is pretrained on OK-VQA and A-OKVQA. Hence, we omit that result from the comparison.}.


\subsection{Main Results}
To demonstrate the effectiveness of our dynamic ensemble of question-answering strategies, we compare our method with state-of-the-art baselines using different types of decision contexts and frozen LLMs. The results are shown in Tab. \ref{tab:results}.

\shortname{} establishes a new state-of-the-art on LLM-based zero-shot KVQA. Compared to previous best scores achieved by frozen LLMs, \shortname{} with Mistral-7B outperforms them with large margins of 2.8\%, 2.6\% and 4.7\% on the OK-VQA, A-OKVQA validation split and test split, respectively. These results strongly demonstrate the effectiveness of our approach.

We also compare against a concurrent method, REPARE \citep{REPARE2024}, which directly prompts an end-to-end trained LVLM. Since REPARE does not translate image to text, it avoids any information loss in the process. Nevertheless, \shortname{} remains highly competitive, outperforming all results of REPARE but that from MiniGPT-4-13B with a much larger LLM, and the difference is merely 0.4\%. The fact that a blind LLM, when prompted properly, is competitive with a well-prompted LVLM at a visual task is surprising to us. As the text-as-visual-representation approach can benefit from advances in LLMs without extensive vision-language alignment training, \shortname{} offers strong practical benefits.

We also compare the time cost of \shortname{} with Img2LLM on Mistral-7B. We randomly select 50 questions from the OKVQA test split and 50 from the A-OKVQA val split. We report the average time spent on each question using two RTX 3090 24G GPUs. Img2LLM takes 1.7 seconds. If we maximize parallelism of LLM calls, \shortname{} takes 15.8 seconds. In completely sequential execution, \shortname{} takes 29.5 seconds. \shortname{} provides another point on the time-accuracy Pareto front and allows a VQA system to trade compute for answer quality when additional compute is available.


\begin{table}[t]
\centering
    \resizebox{\linewidth}{!}{
\begin{tabular}{@{}cccccc|cc@{}}
\toprule
    & \multirow{2}{*}{Cap.} &  & \multirow{2}{*}{SK} &  & \multirow{2}{*}{LK} & OK-VQA & A-OKVQA \\
    &                       &  &                     &  &                     & test   & val     \\ \midrule
\#1 & \checkmark                    &  &                     &  &                     & 46.3   & 44.3    \\
\#2 &                       &  & \checkmark                  &  &                     & 48.1   & 45.9    \\
\#3 &                       &  &                     &  & \checkmark                  & 46.7   & 45.5    \\
\#4 & \checkmark                    &  & \checkmark                  &  &                     & \underline{48.4}   & \underline{47.1}    \\
\#5 & \checkmark                    &  &                     &  & \checkmark                  & 47.1   & 46.8    \\
\#6 & \checkmark                    &  & \checkmark                  &  & \checkmark                  & \textbf{49.2}   & \textbf{47.5}    \\ \bottomrule
\end{tabular}
}
\caption{Ablation study results of diversification with Mistral-7B. Cap. denotes captions. SK and LK are short-form knowledge and long-form knowledge, respectively.}
\label{tab:HRS}
\end{table}

\begin{figure}[t]
    \centering
\includegraphics[width=\linewidth]{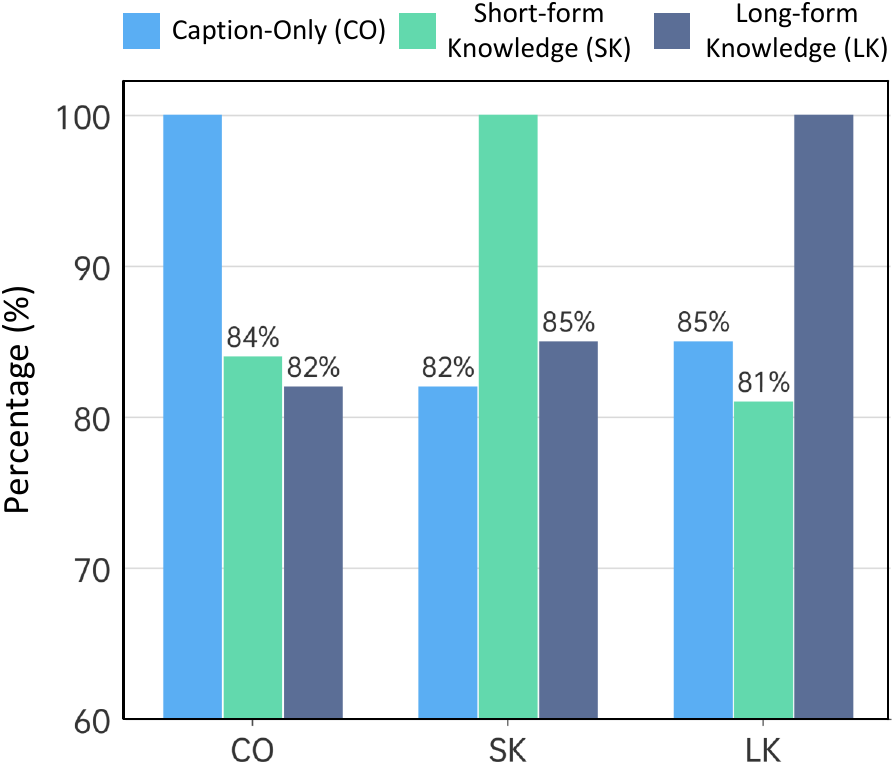}
    \caption{The success rate of the other two strategies on all questions successfully answered by one strategy. 
    In left group, we show the percentage of correctly answered questions by SK and LK out of all the questions that are correctly answered by CO. The other two bar groups are drawn similarly.}
    \label{fig:synergy-of-QA-strategies}
\end{figure}

\begin{table}[t]
\centering
    \resizebox{\linewidth}{!}{
\begin{tabular}{@{}ccccc|ccc@{}}
\toprule
    &  & \multirow{2}{*}{MR} &  & \multirow{2}{*}{AR} & OKVQA &  & A-OKVQA \\
    &  &                     &  &                     & test  &  & val     \\ \midrule
\#1 & \multicolumn{4}{c|}{Rand. Answer Selection}     & 47.4  &  & 45.0    \\ \midrule
\#2 &  &                     &  &                     & 48.0      &  & 46.1        \\
\#3 &  & \checkmark           &  &                     & 48.5  &  & 46.4    \\
\#4 &  &                     &  & \checkmark           & \underline{48.7}  &  & \underline{47.2}    \\
\#5 &  & Rand.               &  & \checkmark           & 48.7  &  & 46.6    \\
\#6 &  & \checkmark           &  & \checkmark           & \textbf{49.2}  &  & \textbf{47.5}    \\ \bottomrule
\end{tabular}
}
\caption{Ablation study of rationalization with Mistral-7B. MR and AR denote mechanistic rationale and automatic rationale, respectively.}
\label{tab:RSF}
\end{table}

\subsection{Ablation Studies}
\label{sec:ablation}
\paragraph{The QA Strategies. } We systematically ablate the three answer strategies, the caption-only, the short-form knowledge, and the long-form knowledge. The outcomes are detailed in Tab. \ref{tab:HRS}. Ablations \#1, \#2, and \#3 utilize each question-answering strategy independently. Ablations \#4 and \#5 utilize partial combinations of the strategies. 

Comparing \#1 with \#2 and \#3, we observe that the short-form knowledge strategy and long-form knowledge strategy achieve better performance than caption-only strategy, demonstrating that generated background knowledge is advantageous in K-VQA. 
After fusing caption-only and short-form knowledge in \#4, the performance on OKVQA and A-OKVQA improves by 2.1\% and 2.8\% over \#1.  This demonstrates the importance of strategy ensemble and the effectiveness of the short-form knowledge strategy. The same phenomenon can be observed in \#1 and \#5. Finally, the full \shortname{} system, \#6, achieves the best performance.

Further, we quantify how much each QA strategy complements others. In Fig. \ref{fig:synergy-of-QA-strategies}, for each QA strategy, we first extract all questions that this strategy can successfully answer by itself. Out of these questions, we show the success rate of the other two strategies. The results indicate that out of the questions answerable by one strategy, 15-19\% cannot be answered by another strategy. Hence, each QA strategy is indispensable  for the success of \shortname.

\begin{figure}[t]
    \centering
\includegraphics[width=\linewidth]{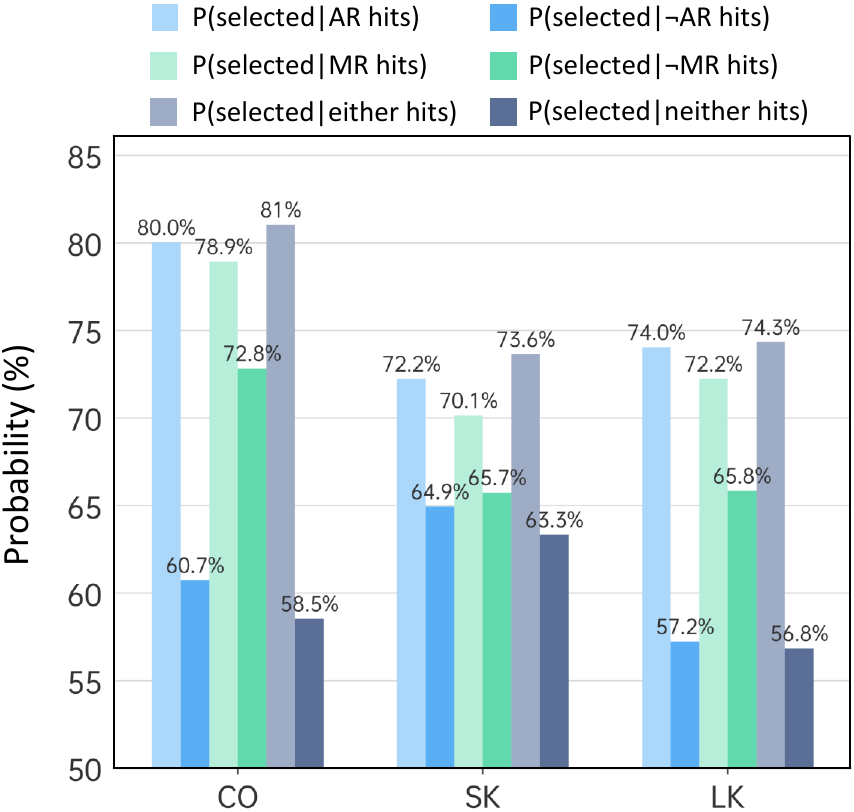}
    \caption{Conditional probabilities of the three answer candidates being selected conditioned on the exact occurrence of the answer in the rationales. 
    There are six different conditions, including the answer candidate occurring in the automatic rationale (AR), the mechanistic rationale (MR), either rationale, and their respective negations. The abbreviations of strategy names are: CO = caption-only, SK = short-form knowledge, and LK = long-form knowledge.}
    \label{fig:rationale-hit-rate}
\end{figure}

\begin{figure*}[t]
\centering
\includegraphics[width=\linewidth]{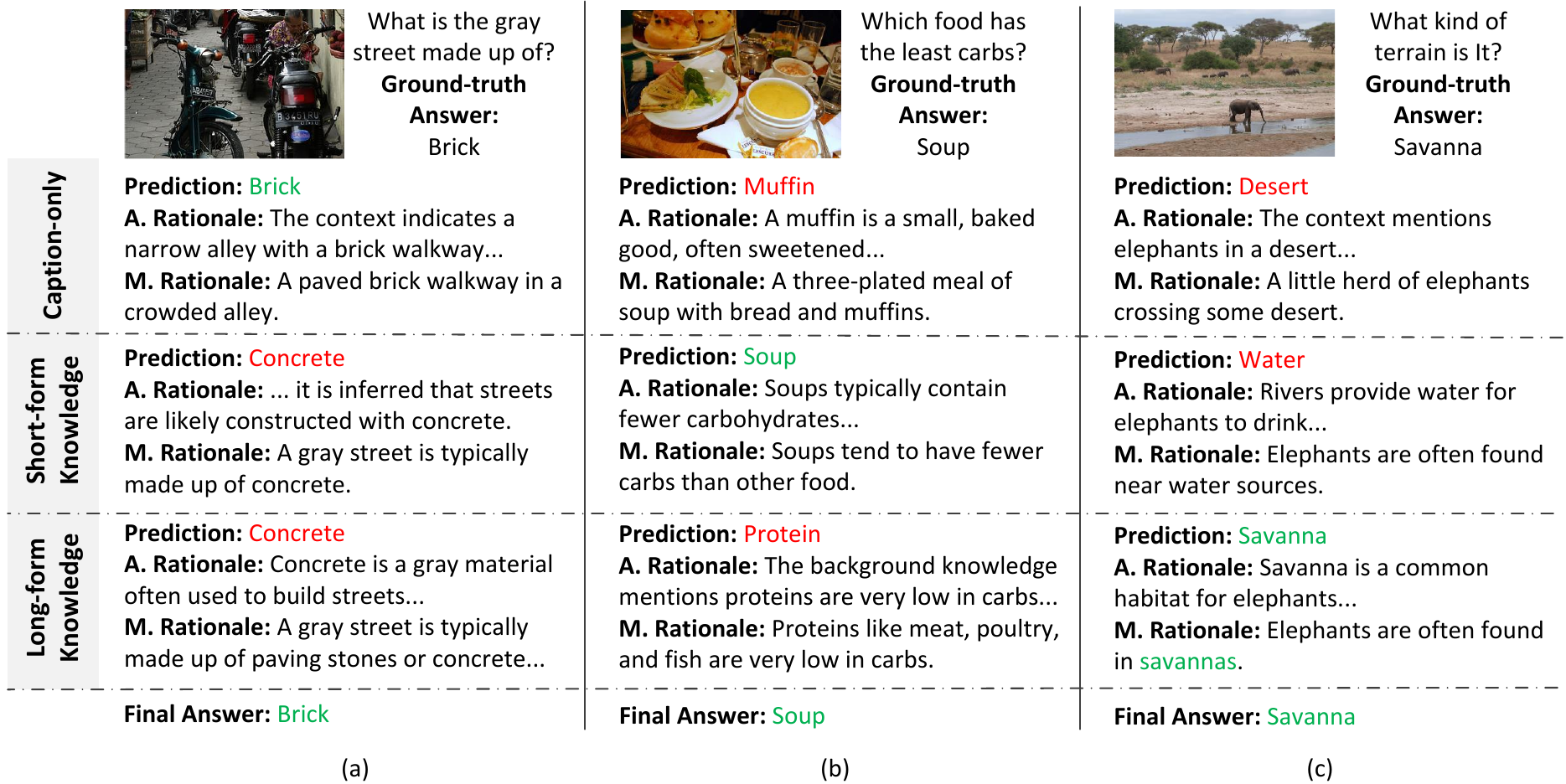}
\caption{Examples from A-OKVQA val split where our method can help model select the correct answer. From top to bottom are the caption-only strategy, the short-form knowledge strategy, and the long-form knowledge strategy. To save space, please refer to Fig. \ref{fig:intro} for the relevant captions and knowledge information.} 
\label{fig:cases}
\end{figure*}
\paragraph{The Rationalization Strategies. } We incrementally integrate automatic and mechanistic rationales into \shortname{}. The results are shown in Tab. \ref{tab:RSF}. The additional baseline (\#1) selects answer candidate randomly without LLM-based answer fusion. The ablation \#2 selects answers using an LLM without rationales. \#5 replaces the mechanistic rationale with a random sentence from the decision context. 

Comparing \#1 with \#2, we observe minor improvements (0.6\% and 1.1\%) from the simplistic answer ensemble without rationales. \#3 and \#4 improve over \#2 by 0.5\% and 0.7\% on OKVQA and by 0.3\% and 1.1\% on A-OKVQA, demonstrating both types of rationales are effective by themselves.  
Adding a random mechanistic rationale to \#4, we obtain ablation \#5, which retains the same performance on OKVQA but decreases by 0.6\% on A-OKVQA. This highlights the importance of selecting proper mechanistic rationales, as wrong rationales are harmful. Most interestingly, leveraging both rationales, \#6 attains the best performance, +1.2\% / +1.4\% over \#2. Notably, the improvements are exactly the sum of the improvements attained by the individual rationales (\#3 and \#4), suggesting the two rationales are perfectly complementary and the ensemble over rationales is effective.

Finally, with Fig. \ref{fig:rationale-hit-rate}, we analyze the relation between the rationale surface form containing exactly the  answer candidate, which we refer to as a ``hit'', and the candidate being selected by the ensemble phase LLM. Similar to the finding of \citet{pnpvqa}, we observe a positive correlation between the two events. This suggests that the surface form of the rationales plays a role in the decisions of the answer-selecting LLM. 
The largest differential is with caption-only, where having the answer in either rationale increases the selection probability by 22.5\%\footnote{P(selected $\mid$ either hits)$-$P(selected $\mid$ neither hits)=22.5\%}. In contrast, having the candidate in the rationales under SK or LK is not as important (probability differences of 17.5-10.3\%). A possible reason is that the LLM conducts more complex operations on questions requiring external knowledge than simple surface-form matching.

\subsection{Case Study}
We show some example K-VQA questions in Fig. \ref{fig:cases}. To conserve space, we do not repeat the corresponding captions and knowledge details, available in Fig. \ref{fig:intro}. In strategy ensemble, we provide answer candidates with corresponding rationales and caption-generated QA pairs in the prompt. More examples can be found in Fig. \ref{fig:cases_appendix} of the Appendix \ref{sec:appendix}.  

In example (a), since ``Brick walkway'' appears in the captions, the answer from caption-only strategy is ``Brick''. However, as concrete is a more common street material, the answers from short-form and long-form knowledge are both ``Concrete''. Still, the model selects "Brick" as the final answer, possibly due to the strength of the rationale. In example (b), the answer from the caption-only strategy is "Muffin", but the its rationales are not proper explanations. The answer from the short-form knowledge strategy is correct and has sensible rationales, and it is selected as the final answer. In example (c), desert and savanna are both possible answers, which are hard to distinguish without visual information. The model may have been informed by the rationales from the second strategy mentioning rivers, which are more common in a savanna than in a desert. 

\section{Conclusion}
 We propose \shortname, which ensembles several QA strategies for knowledge-based VQA. \shortname{} first generates three types of decision contexts: image captions, short-form knowledge and long-form knowledge, and answers the question from each decision context separately. After that, \shortname{} generates two types of rationales for each answer. An ensemble LLM selects the best answer from the rationales. \shortname{} achieves state-of-the-art results on OK-VQA among comparable methods and showcases LLM-informed, rationale-based ensemble as an effective VQA method. \shortname{} allows potential trade-off between compute and answer quality and may contribute to research on scaling LLMs with inference-time compute, a research direction pioneered by OpenAI-o1 \cite{OpenAI-o1}.  

\section*{Acknowledgments}
We gratefully acknowledge the support by the Nanyang Associate Professorship and the National Research Foundation Fellowship (NRF-NRFF13-2021-0006), Singapore. Any opinions, findings, conclusions, or
recommendations expressed in this material are
those of the authors and do not reflect the views of
the funding agencies.

\section*{Limitations}
In this paper, we generate three different types of decision contexts for each question, including captions, short-form knowledge, and long-form knowledge, where both types of knowledge are generated based on captions. However, in the process of converting images to captions, some visual details are inevitably lost, resulting in inaccurate captions. Therefore, the knowledge generated based on these captions may also be inaccurate. These inaccurate decision contexts may lead to incorrect answers. In extreme cases, if all answer candidates are wrong, our method will fail. Hence, improving the quality of decision contexts is an important direction for future research. Moreover, our method requires multiple calls to the LLM for inference, which allows inference-time scaling up of model capabilities but may take too long. Reducing the running time could be a direction for future research. 

Although our work has achieved good results, it also inherits any existing biases of LLMs and their training datasets. Future work can focus on addressing these issues.

\bibliography{custom}

\appendix
\section{Experimental Details}
We use the instruction-tuned LLM in generation of the knowledge text and the automatic rationale, and the LLM before instruction tuning in generation of the answer and the mechanistic rationale. 
For the standard version of the LLM, we use a greedy decoding strategy. For the instruction-tuned version of the LLM, we use top-k sampling, where k is set to 50. We set the temperature, the length penalty and the repetition penalty all to 1.0, and set the diversity penalty to 0. 

\section{Prompt Templates}
\label{sec:appendix}
In this section, we demonstrate prompt templates for generating answer candidate, background knowledge, automatic rationale, and question-answering strategy fusion, as shown in Tab. \ref{tab:PromAns}, Tab. \ref{tab:PromKnow}, Tab. \ref{tab:PromAR} and Tab. \ref{tab:PromFus}, respectively. We use the same prompt templates across different datasets and models.

\begin{table*}[]
\centering
    \resizebox{0.8\textwidth}{!}{
\begin{tabular}{@{}lll@{}}
\toprule
\multicolumn{3}{l}{Prompt Template for Generating Answer Candidate}                          \\ \midrule
\multicolumn{3}{l}{Please answer questions according to the given context.}                   \\
\multicolumn{3}{l}{Context: \textit{$<$Captions$>$}$/$\textit{$<$Short-form Knowledge$>$}$/$\textit{$<$Long-form Knowledge$>$}} \\
\multicolumn{3}{l}{\textit{$<$Question-Answer Pairs$>$}}                                            \\
\multicolumn{3}{l}{Question: \textit{$<$Question$>$}}                                               \\
\multicolumn{3}{l}{Answer:}                                                                   \\ \bottomrule
\end{tabular}
}
\caption{Prompt template for generating answer candidate. Choose one from three decision contexts.}
\label{tab:PromAns}
\end{table*}

\begin{table*}[]
\centering
    \resizebox{0.95\textwidth}{!}{
\begin{tabular}{@{}lll@{}}
\toprule
\multicolumn{3}{l}{Prompt Template for Generating Background Knowledge}                                        \\ \midrule
\multicolumn{3}{l}{\textbf{User:} You are going to answer questions according to the context: \textit{$<$Captions$>$}}                                                \\
\multicolumn{3}{l}{\textbf{Assistant:} Ok, please go ahead and ask your questions.}       \\
\multicolumn{3}{l}{\textit{$<$Question-Answer Pairs$>$}}       \\
\multicolumn{3}{l}{\textbf{User:} \textit{$<$Question$>$}}                                                \\
\multicolumn{3}{l}{\textbf{Assistant:} I don't have enough knowledge to answer this question.}                          \\
\multicolumn{3}{l}{\textbf{User:} Please provide background knowledge related to this question (in a single sentence).} \\ \bottomrule
\end{tabular}
}
\caption{Prompt template for generating background knowledge. We assume the model cannot answer the question using a caption-only strategy and ask it to generate relevant background knowledge.}
\label{tab:PromKnow}
\end{table*}

\begin{table*}[]
\centering
    \resizebox{0.8\textwidth}{!}{
\begin{tabular}{@{}lll@{}}
\toprule
\multicolumn{3}{l}{Prompt Template for Generating Automatic Rationale}                                        \\ \midrule
\multicolumn{3}{l}{\textbf{User:} You are going to answer questions according to the context:} \\
\textit{$<$Captions$>$}$/$\textit{$<$Short-form Knowledge$>$}$/$\textit{$<$Long-form Knowledge$>$}                                                \\
\multicolumn{3}{l}{\textbf{Assistant:} Ok, please go ahead and ask your questions.}       \\
\multicolumn{3}{l}{\textit{$<$Question-Answer Pairs$>$}}       \\
\multicolumn{3}{l}{\textbf{User:} \textit{$<$Question$>$}}                                                \\
\multicolumn{3}{l}{\textbf{Assistant:} \textit{$<$Answer$>$}}                          \\
\multicolumn{3}{l}{\textbf{User:} Please explain the reasoning behind your answer in a single sentence.} \\ \bottomrule
\end{tabular}
}
\caption{Prompt template for generating automatic rationale. Choose one from three decision contexts. We provide the answer candidate of corresponding strategy and ask the model to explain the reasoning behind it.}
\label{tab:PromAR}
\end{table*}

\begin{table*}[]
\centering
    \resizebox{0.8\textwidth}{!}{
\begin{tabular}{@{}lll@{}}
\toprule
\multicolumn{3}{l}{Prompt Template for Strategy Fusion}                                \\ \midrule
\multicolumn{3}{l}{Please answer the question based on the most reasonable rationale.} \\
\multicolumn{3}{l}{Rationales:}                                                        \\
\multicolumn{3}{l}{1.$<A.\,Rationale\,of\,Answer_1>.\,<M.\,Rationale\,of\,Answer_1>$}                            \\
\multicolumn{3}{l}{2.$<A.\,Rationale\,of\,Answer_2>.\,<M.\,Rationale\,of\,Answer_2>$}                            \\
\multicolumn{3}{l}{3.$<A.\,Rationale\,of\,Answer_3>.\,<M.\,Rationale\,of\,Answer_3>$}                            \\
\multicolumn{3}{l}{$<Question-Answer\,Pairs>$}                                        \\
\multicolumn{3}{l}{Question: $<Question>\,<Answer_1>$ or $<Answer_2>$ or $<Answer_3>$?}                                                     \\
\multicolumn{3}{l}{Answer:}                                                            \\ \bottomrule
\end{tabular}
}
\caption{Prompt template for ensemble. We first provide rationales for each answer, then ask the model to select a final answer from the answer candidates based on these rationales.}
\label{tab:PromFus}
\end{table*}

\section{Interactions Between Decision Contexts}
To explore the effects of interactions between decision contexts, we conduct additional experiments using Mistral-7B by concatenating different decision contexts in one prompt. The result is 41.7\% (44.7\% resp.)  on A-OKVQA val split (OKVQA test split resp.), which is much lower than 47.5\% (49.2\% resp.) of our method. Consistent with our observation, \citet{KGVQA} also concatenate captions and long-form knowledge in KGenVQA, and the result is 39.1\% (45.4\% resp.). We hypothesize that concatenating different decision contexts into a single prompt can introduce contradictory or redundant information, making it  difficult for the model to effective utilize the information in the context to answer questions.

\section{Ensemble of Different Caption-only Strategies}
We conduct experiments using three different image captioning models, BLIP \citep{blip}, InstructBLIP \citep{InstructBLIP-2023} and OFA \citep{wang2022ofa} to generate different captions. We replace the short-form and long-form knowledge in our method with the captions generated by InstructBLIP and OFA. The result with Mistral-7B on A-OKVQA val split (OKVQA test split resp.) is 44.5\% (46.7\% resp.), much lower than 47.5\% (49.2\% resp.) from our method.

\section{Examples}
\label{sec:appendixB}
\begin{figure*}[t]
\centering
\includegraphics[width=\linewidth]{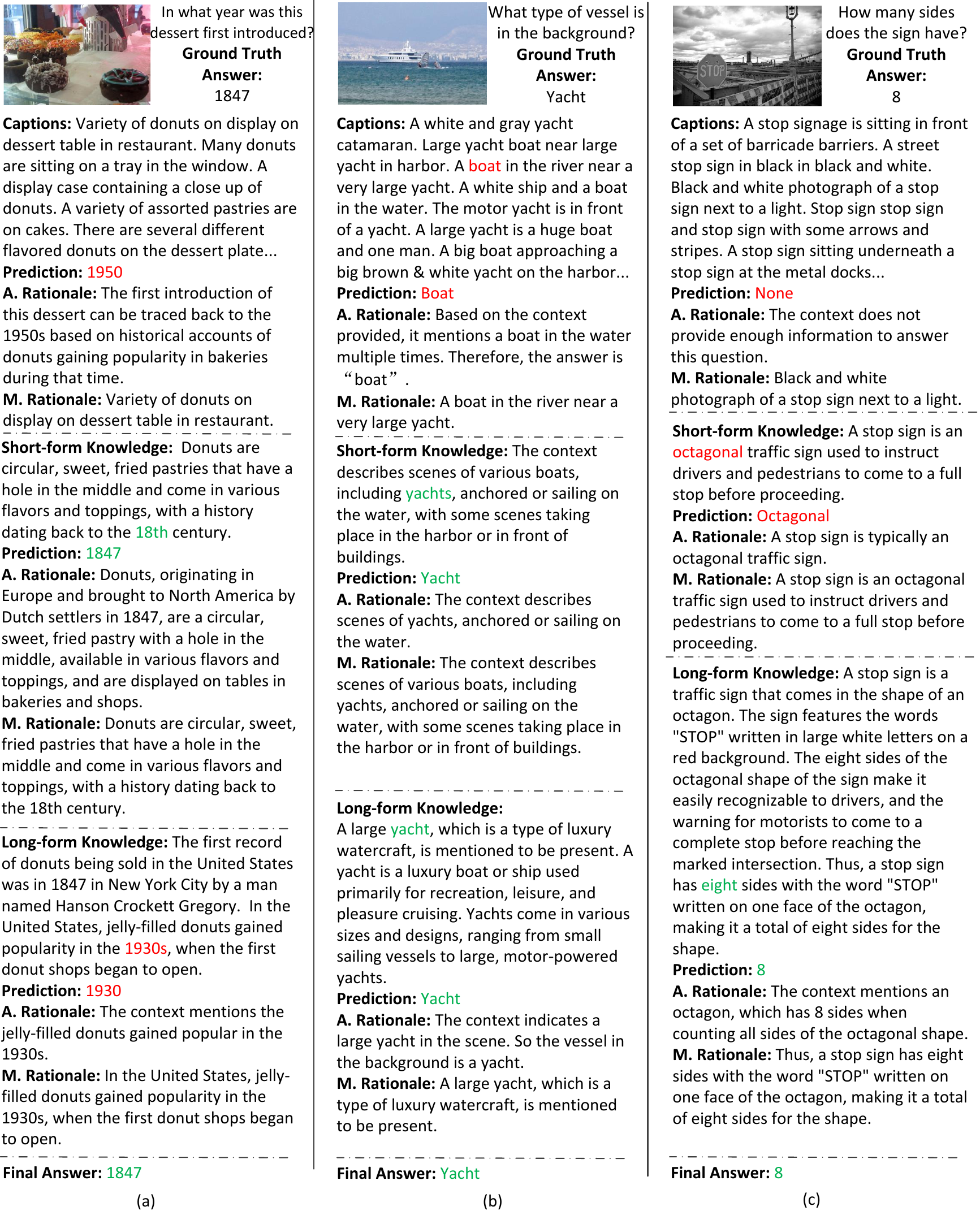}
\caption{Examples from OKVQA where our method can help model select the correct answer. From top to bottom are the caption-only strategy, the short-form knowledge strategy, and the long-form knowledge strategy. Incorrect answers are marked with \textcolor{red}{red} and correct answers are in \textcolor{LimeGreen}{green}. } 
\label{fig:cases_appendix}
\end{figure*}

In this section, we display more examples from OK-VQA where \shortname{} can help model select the correct answer, as shown in Fig \ref{fig:cases_appendix}.

Understanding why the LLM selects certain answers is challenging for two main reasons. First, as recent research \citep{paradox} indicates, LLMs exhibit behaviors drastically different from humans and may not fully comprehend the outputs they generate. That is, even if the LLM generates the right answer, it may not be able to explain why it selects a particular answer, or do other things that we expect a human who make the right choice would be able to do. Second, without ground-truth annotations for errors in the rationales, it is difficult to quantitatively assess if LLMs correctly detect illogical rationales or factual errors.

Despite these challenges, in Fig. \ref{fig:rationale-hit-rate}, we analyze the relation between the rationale surface form containing exactly the answer candidate and the candidate being selected by the answer-selecting ensemble LLM. The positive correlation between the two events suggests that the surface form of the rationales plays a role in the decision of the answer-selecting LLM. 

In addition, we provide three examples by asking the LLM to explain its choice to illustrate how the LLM selects answers. We use ``Can you explain briefly how you select the final answer based on the three answer candidates and corresponding rationales?'' as the instruction, and providing the final answer and three answer candidates with their rationales in the prompt. For question (a) in Fig. \ref{fig:cases_appendix}, the output is ``1847 is the final answer because it marks the historical introduction of donuts to North America, supported by corresponding rationale.''
We speculate that the rationale behind the answer "1847" is more relevant to the question regarding the "first introduced" date, so the model selects "1847" as the final answer. 

For question (b), the output is ``The final answer `Yacht' was selected because multiple rationales consistently describe a large luxury vessel in the scene, and the term `yacht' aligns more specifically with that description than the more general term `boat'.'' For question (c), the output is ``I selected the final answer `8' because it provides the most precise and direct explanation by identifying the number of sides (8) of an octagonal stop sign.'' 

\begin{figure*}[t]
\centering
\includegraphics[width=\linewidth]{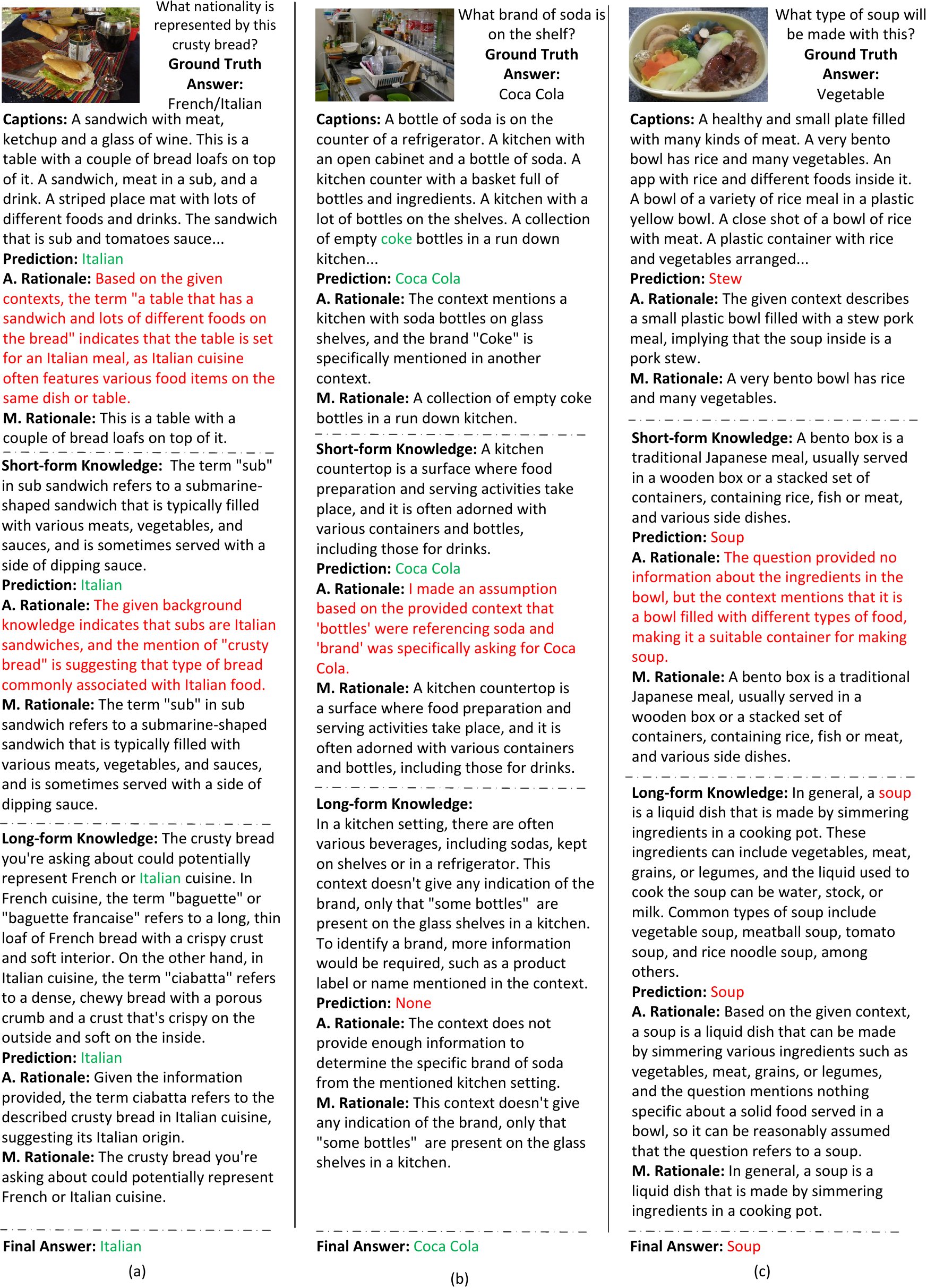}
\caption{Examples from OKVQA where certain rationales are wrong. The wrong rationales and answers are marked in \textcolor{red}{red}, and correct answers are in \textcolor{LimeGreen}{green}.} 
\label{fig:cases2_appendix}
\end{figure*}

We also provide some examples containing wrong rationales in Fig. \ref{fig:cases2_appendix}.

\end{document}